\documentclass[runningheads]{llncs}
\usepackage{graphicx}
\usepackage{amsmath,amsfonts}
\usepackage{amssymb}
\usepackage{xfrac}
\usepackage{booktabs}
\usepackage{multirow}
\usepackage{pifont}
\usepackage{graphicx}
\usepackage[utf8]{inputenc} 
\usepackage{hyperref}       
\usepackage{url}            
\usepackage{booktabs}       
\usepackage{amsfonts}       
\usepackage{nicefrac}       
\usepackage{microtype}      
\usepackage{amsmath}
\usepackage{mathrsfs}
\usepackage{amssymb}
\usepackage{multirow}
\usepackage{multicolrule}
\usepackage{booktabs}
\usepackage{algorithm}
\usepackage{algorithmic}
\usepackage{times}
\usepackage{helvet}
\usepackage{courier}
\usepackage{adjustbox}
\usepackage[font=normal]{subfig}
\usepackage[misc]{ifsym}

\newcommand{\tabincell}[2]{\begin{tabular}{@{}#1@{}}#2\end{tabular}}
\makeatletter
\newcommand{\shorteq}{%
  \settowidth{\@tempdima}{-}
  \resizebox{\@tempdima}{\height}{=}%
}
\makeatletter
\newcommand{\printfnsymbol}[1]{\textsuperscript{\@fnsymbol{#1}}}
\makeatother
\newcommand{\PreserveBackslash}[1]{\let\temp=\\#1\let\\=\temp}
\newcommand*{\affaddr}[1]{#1} 
\newcommand*{\affmark}[1][*]{\textsuperscript{#1}}

\institute{}
\begin{document}
\title{Generalized Organ Segmentation by Imitating One-shot Reasoning using Anatomical Correlation}
%
%
\author{Hong-Yu Zhou\thanks{First three authors contributed equally. Work done at Tencent}\affmark[1] \and
Hualuo Liu\printfnsymbol{1}\affmark[2] \and Shilei Cao\printfnsymbol{1}\affmark[2] \and Dong Wei\thanks{Corresponding author}\affmark[2] \and Chixiang Lu\affmark[3] \and Yizhou Yu\affmark[1] \and Kai Ma\affmark[2] \and Yefeng Zheng\affmark[2] \\
\affaddr{\affmark[1]The University of Hong Kong}\\
\affaddr{\affmark[2]Tencent}\\
\affaddr{\affmark[3]Huazhong University of Science and Technology}\\
\email{\{whuzhouhongyu, eliasslcao, luchixiang\}@gmail.com}\\
\email{lhl18@mails.jlu.edu.cn}\\
\email{\{donwei,kylekma,yefengzheng\}@tencent.com}\\
\email{yizhouy@acm.org}
} 
%
%
%
\maketitle  
\begin{abstract}
Learning by imitation is one of the most significant abilities of human beings and plays a vital role in human's computational neural system. In medical image analysis, given several exemplars (anchors), experienced radiologist has the ability to delineate unfamiliar organs by imitating the reasoning process learned from existing types of organs. Inspired by this observation, we propose \emph{OrganNet} which learns a generalized organ concept from a set of annotated organ classes and then transfer this concept to unseen classes. In this paper, we show that such process can be integrated into the one-shot segmentation task which is a very challenging but meaningful topic. We propose pyramid reasoning modules (PRMs) to model the anatomical correlation between anchor and target volumes. In practice, the proposed module first computes a correlation matrix between target and anchor computerized tomography (CT) volumes. Then, this matrix is used to transform the feature representations of both anchor volume and its segmentation mask. Finally, OrganNet learns to fuse the representations from various inputs and predicts segmentation results for target volume. Extensive experiments show that OrganNet can effectively resist the wide variations in organ morphology and produce state-of-the-art results in one-shot segmentation task. Moreover, even when compared with fully-supervised segmentation models, OrganNet is still able to produce satisfying segmentation results.

\keywords{One-shot Learning \and Image Segmentation \and Anatomical Similarity.}
\end{abstract}
\section{Introduction}
Organ segmentation has wide applications in disease diagnosis, treatment planning, intervention, radiation therapy and other clinical workflows~\cite{gibson2018automatic}. Thanks to deep learning, remarkable progress has been achieved in organ segmentation tasks. However, existing deep models usually focus on a predefined set of organs given abundant annotations for training (e.g., heart, aorta, trachea, and esophagus in \cite{trullo2019multiorgan}) but fail to generalize to unseen abdominal organ with only limited annotations (in the extreme case, only one annotation is available), which limits their clinical usage in practice.

\begin{figure}[t!]
	\setlength{\belowcaptionskip}{-5pt}
	\begin{center}
		\includegraphics[width=0.55\columnwidth]{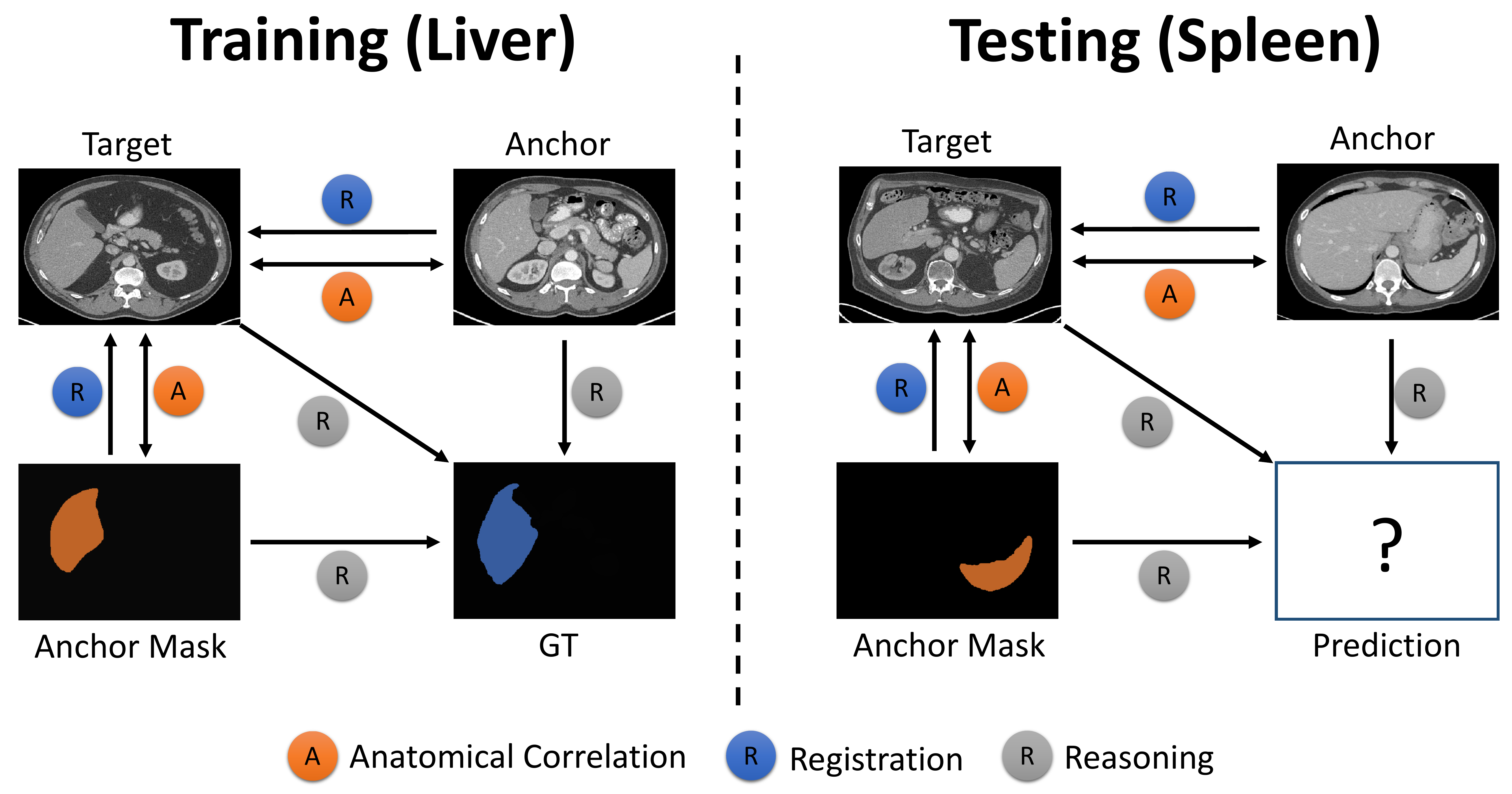}
	\end{center}
	\caption{Conceptual overview of proposed one-shot reasoning process. During the training stage, given a target image and paired anchor image with its annotation mask, we first conduct registration between anchor and target images. Then, the proposed reasoning module tries to model the anatomical correlation between anchor and target images. The correlation matrix is further used to transform the feature representations of both anchor image and its corresponding mask, which are jointly learned with target image features to produce final predictions. We argue that the generalized organ concept can be learned from a set of sufficiently annotated organs in such process (left). In the test stage, the learned knowledge generalize to unseen abdominal organs whose segmentation results can be easily inferred (right). Note that in practice, the inputs are 3D volumes instead of 2D images.}
	\label{fig_illustration}
	\vspace{-8pt}
\end{figure}

A potential solution to this problem is the one-shot segmentation methodology~\cite{zhao2019data,wang2020lt}, which attempts to learn knowledge from only one labeled sample. Nonetheless, these approaches lack the ability to handle large variations among different organ types and thus cannot be directly applied to one-shot organ segmentation. On the other hand, we find that human radiologists naturally maintain the ability to effectively learn unfamiliar organ concepts given limited annotated data, which we think can be attributed to their usage of of anatomical similarity to segment both seen and unseen organs. Meanwhile, it is also a reasonable setting to transfer the learned knowledge from richly annotated organs to less annotated ones. Inspired by these observations, we propose a new one-shot segmentation paradigm where we assume that a \emph{generalized} organ concept can be learned from a set of sufficiently annotated organs and applied for effective one-shot learning to segment some previously unseen abdominal organs. We illustrate the proposed method in Fig. \ref{fig_illustration} where we call it \textsc{one-shot reasoning} as it imitates the reasoning process of human radiologists by making use of anatomical similarity. 

Anatomical similarity have been widely used in medical image segmentation \cite{dinsdale2019spatial,liang2019comparenet}. Compared to these methods, our work mainly exploits using anatomical similarity within each one-shot pair of images to perform reasoning. In this sense, our work provides a new segmentation paradigm by utilizing learned organ priors, with a focus on the one-shot scenario. In this paper, we propose OrganNet to implement the concept of generalized organ learning for one-shot organ segmentation in medical images. Our contributions can be summarized as follows:
\begin{enumerate}
	\item We propose a new organ segmentation paradigm which learns a generalized organ concept from seen organ classes and then generalize to unseen classes using one-shot pairs as supervision.
	\item A reasoning module is developed to exploit the anatomical correlation between adjacent spatial regions of anchor and target computerized tomography (CT) volumes, which can be utilized to enhance the representations of anchor volume and its segmentation annotation.
	\item We introduce OrganNet, which includes two additional encoders to basic 3D U-Net architecture to jointly learn representations from target volume, anchor volume and its corresponding segmentation mask.
	\item We conduct comprehensive experiments to evaluate OrganNet. The experimental results on both organ and non-organ segmentation tasks demonstrate the effectiveness of OrganNet.
\end{enumerate}

\section{Related Work}
Utilizing the anatomical correlation is one of the key designs of proposed OrganNet.
In this section, we first review existing works related to the utilization of anatomical priors and then list the most related works in one-shot medical segmentation.

\subsubsection{Anatomical correlation in medical image segmentation.} A large body of literature \cite{bentaieb2016topology,ravishankar2017joint,ravishankar2017learning} exploits anatomical correlation for medical image segmentation within the deep learning framework. The anatomical correlation also serves as the foundation of atlas-based segmentation \cite{iglesias2015multi,liang2019comparenet}, where one or several labeled reference images (i.e., atlases) are non-rigidly registered to a target image based on the anatomical similarity, and the labels of the atlases are propagated to the target image as the segmentation output. Different from these methods, OrganNet has the ability to learn anatomical similarity between images by employing the reasoning process. With this design, the proposed approach is able to learn a generalized organ concept for one-shot organ segmentation.


\subsubsection{One-shot medical segmentation.} Zhao \textit{et al.} \cite{zhao2019data} presented an automated data augmentation method for synthesizing labeled medical images on resonance imaging (MRI) brain scans. However, DataAug is strictly restricted to segmenting objects when only small changes exist between anchor and target images. Based on DataAug, Dalca \textit{et al.} \cite{dalca2019unsupervised} further extended it to an alternative strategy that combines a conventional probabilistic atlas-based segmentation with deep learning. Roy \textit{et al.} \cite{roy2020squeeze} proposed a few-shot volumetric segmenter by optimally pairing a few slices of the support volume to all the slices of the query volume. Similar to DataAug, Wang \textit{et al.} \cite{wang2020lt} introduced cycle consistency to learn reversible voxel-wise correspondences for one-shot medical image segmentation. Lu \textit{et al.} \cite{lu2020learning} proposed a one-shot anatomy segmentor which is based on a naturally built-in human-in-the-loop mechanism. Different from these approaches, this paper focuses on a more realistic setting: we use richly annotated organs to assist the segmentation of less annotated ones. Also, our method is partially related to the siamese learning \cite{zhou2020comparing,zhou2017sunrise,bertinetto2016fully,zhou2018semi} which often takes a pair of images as inputs.

\section{Proposed Method}
In this section, we introduce OrganNet, which is able to perform one-shot reasoning by exploiting anatomical correlation within input pairs. Particularly, we propose to explore such correlation from multiple scales where we use different sizes of neighbour regions as shown in Fig. \ref{framework}. More details will be presented in the following.

Before we send anchor and target images (in practice, CT volumes) to OrganNet, it is suggested that registration should be conducted in order to align their image space. Since most of data come from abdomen, we apply  DEEDS (DEnsE Displacement Sampling) \cite{heinrich2013mrf} as it yielded the best performance in most CT-based datasets \cite{xu2016evaluation}. Note that the organ mask is also aligned according to its corresponding anchor image.

\subsection{One-shot Reasoning using OrganNet}
\label{sec:method}
\begin{figure*}[t]
	\centering
	\setlength{\belowcaptionskip}{-5pt}
	\includegraphics[width=0.65\linewidth]{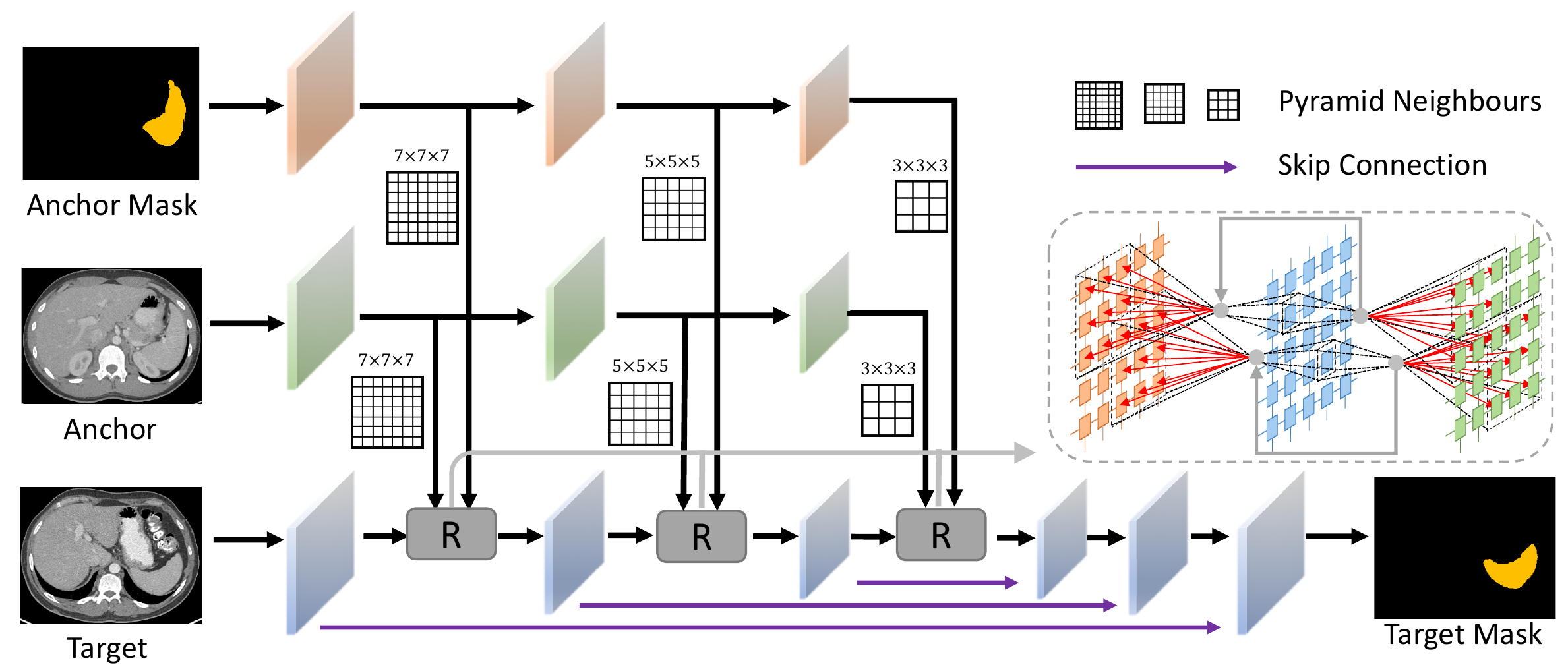}
	\caption{Network structure of proposed OrganNet. It augments the classic U-Net structure with two additional encoders to learn to perform one-shot reasoning from an anchor image and its corresponding segmentation annotation. Pyramid reasoning modules (PRMs, denoted as \textbf{R}) are proposed to model the anatomical similarity between target and anchor volumes. Note that all operations, and the input and output of OrganNet are 3D; here we use 2D slices of the 3D volumes for the purpose of idea illustration.}
	\label{framework}
\end{figure*}

As shown in Fig. \ref{framework}, OrganNet has three encoders which learn representations for target volume, anchor volume and anchor mask, respectively. Moreover, we propose pyramid reasoning modules (PRMs) to connect different encoders which decouples OrganNet into two parts: one part for learning representations from anchor volume and its paired organ mask (top two branches) and the other for exploiting anatomical similarity between the anchor and the target volumes (bottom two branches). We argue that the motivation behind can be summarized as: the generalized organ concept can be learned from a set of sufficiently annotated organs, and then generalize to previously unseen abdominal organs by utilizing only a single one-shot pair.

The OrganNet is built upon the classic 3D U-Net structure \cite{cciccek20163d} and we extend it to a tri-encoder version to include extra supervision from anchor image and its annotation. In practice, we ddesign OrganNet to be light-weight in order to alleviate the overfitting problem caused by small datasets in medical image analysis. Specifically, for each layer, we only employ two bottlenecks for both encoders and only one bottleneck for the decoder branch. Since all operations are in 3D, we use a relatively small number of channels for each convolutional kernel to reduce computational cost.

Imitation is a powerful cognitive mechanism that humans use to make inferences and learn new abstractions \cite{gentner1997reasoning}. We propose to model the anatomical similarity between images to provide strong knowledge prior for learning one-shot segmentation. However, the variations in organ location, shape, size, and appearance among individuals can be regarded as an obstacle for a network to make reasonable decisions. To address these challenges, we propose PRMs to effectively encapsulate representations from multiple encoders.

\subsection{Pyramid Reasoning Modules}
These modules are designed to address the situation in which the organ morphologies and structures of target and anchor volumes show different levels of variations. Since features in different feature pyramids capture multi-scale information, we propose to aggregate information at each pyramid level with a reasoning function. To account for the displacement between the two images, large sizes of neighbour regions are employed in shallow layers, whereas small region sizes are employed in deep layers. The underlying reason of such allocation is that the receptive fields of shallow layers are smaller than those of deep ones. Concretely, we first compute the correlation matrix between feature maps of target and anchor volumes. Then, we apply this matrix to transform feature representations of anchor input and its segmentation mask, respectively. Finally, we concatenate representations of three inputs and treat them as the input to next layer.
\begin{figure}[t]
	\centering
	\setlength{\belowcaptionskip}{-5pt}
	\subfloat[][Overview of the reasoning module]{
	\includegraphics[width=0.53\columnwidth]{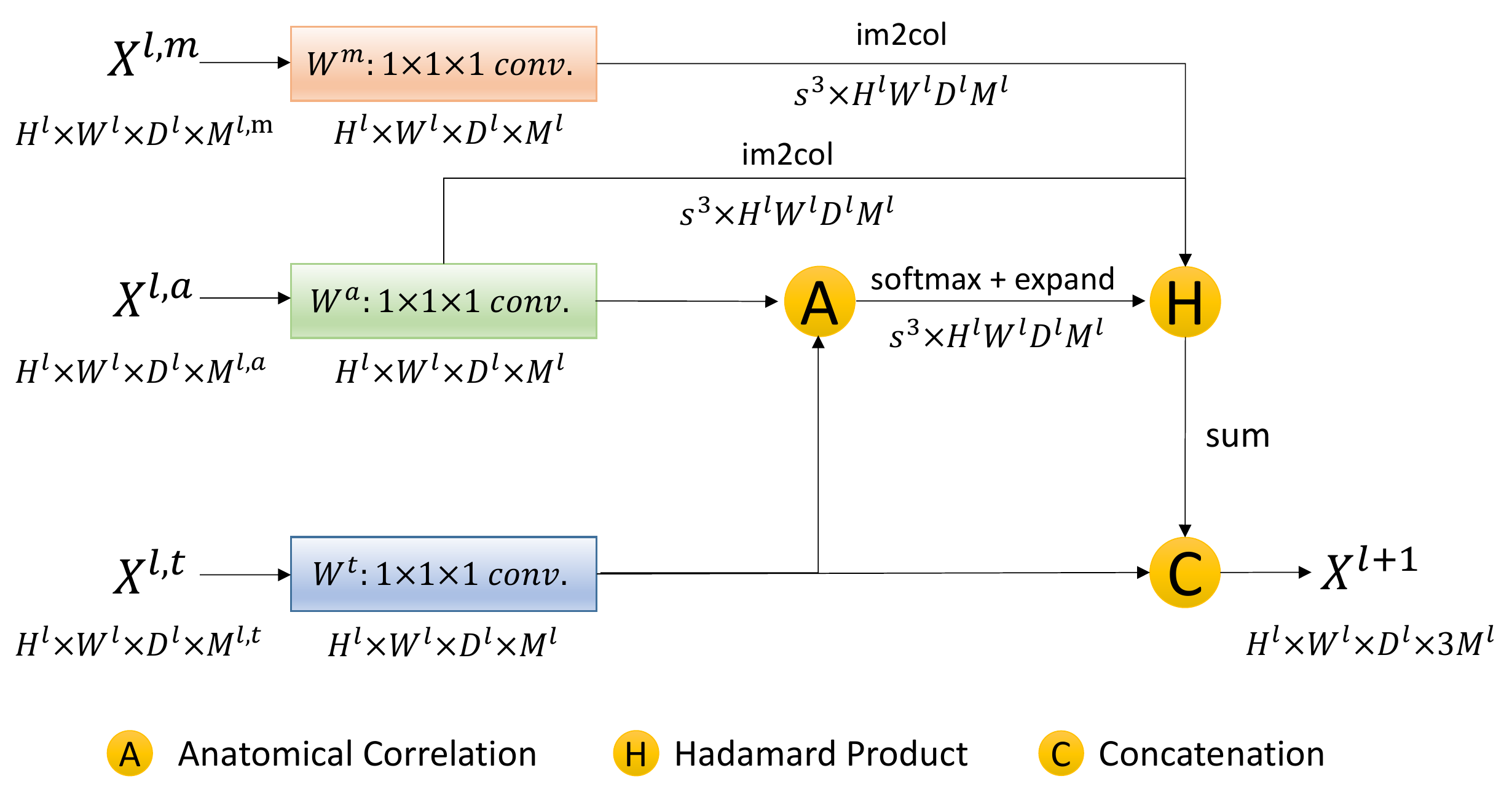}}
	\ \ 
	\subfloat[][2D illustration]{
	\includegraphics[width=0.41\columnwidth]{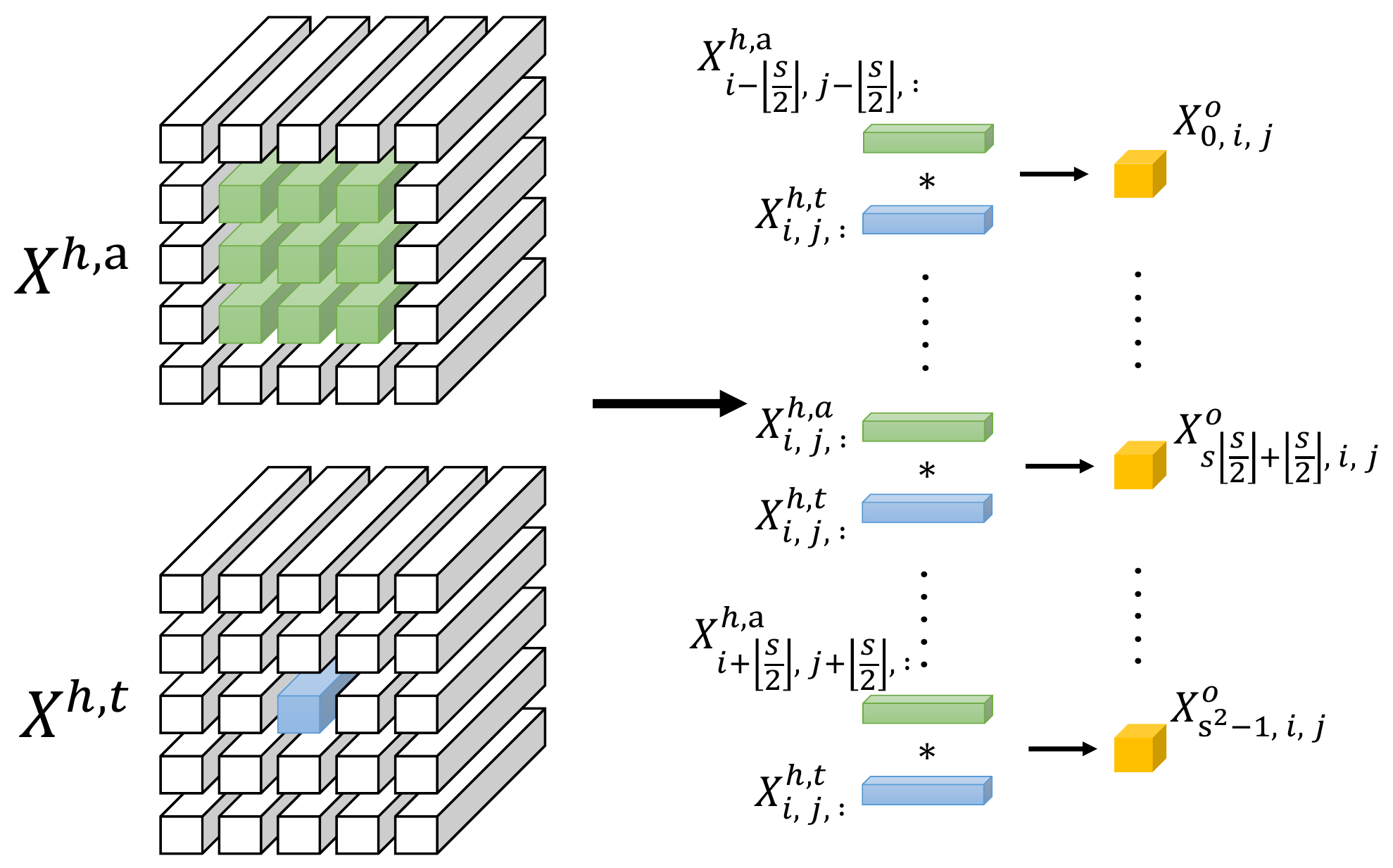}}
	\caption{Main architecture of the reasoning module. (a) $X^{l,t}$, $X^{l,a}$ and $X^{l,m}$ represent layer $l$'s input tensors for target volume, anchor volume and anchor mask, respectively. We apply \emph{softmax} to tensor's first dimension where \emph{sum} operation is also conducted after Hadamard product. In (b), we provide a simplified 2D illustration of the reasoning module.}
	\label{reasoning}
\end{figure}

As shown in Fig. \ref{reasoning}, at layer $l$, each reasoning module has three input tensors $X^{l,t}$, $X^{l,a}$ and $X^{l,m}$, corresponding to target volume, anchor volume and anchor mask, respectively. We first apply three $3\times 3\times 3$ convolutional operators to above three input tensors in order to normalize them to the same scale. The outputs of such operations are $X^{h,t}$, $X^{h,a}$ and $X^{h,m}$, and the size of each is $H^l\times W^l\times D^l\times M^l $. 

To model the anatomical correlation between $X^{h,t}$ and $X^{h,a}$, we apply inner product operation to neighbour regions from both tensors. Particularly, given a specific vector $X^{h,t}_{i,j,k,:}$ in tensor $X^{h,t}$ with $0 \leq i<H^l$, $0 \leq j<W^l$ and $0\leq k<D^l$, we compute the inner product based on its neighbour regions in $X^{h,a}$ which can be summarized as:
\begin{align}
	\footnotesize
	X^{o}_{n,i,j,k}=(X^{h,a}_{i+i',j+j',k+k',:})^T X^{h,t}_{i,j,k,:}
\end{align}
where $\{i', j', k'\} \in \{-\lfloor \frac{s}{2} \rfloor,..., \lfloor \frac{s}{2} \rfloor \}$. $s$ stands for the size of neighbour region which changes with layer depth. And $n$ stands for the total index which can be computed as:
\begin{align}
	n = (i'+\lfloor\frac{s}{2}\rfloor) + (j'+\lfloor\frac{s}{2}\rfloor)\times s + (k'+\lfloor\frac{s}{2}\rfloor)\times s^2.
\end{align}
$X^o \in \mathcal{R}^{s^3\times H^l\times W^l\times D^l}$. For better understanding,, we provide an illustration of the computation process of anatomical correlation in 2D network, which can be found in the supplementary material.

Now $X^o$ represents the anatomical similarity between $X^{h,t}$ and $X^{h,a}$. Then, we apply \emph{softmax} normalization along with the first dimension of $X^o$ and \emph{expand} its dimension to $s^3\times H^l W^l D^l M^l$, which can be summarized as:
\begin{align}
	X^w = {{expand}}({{softmax}}(X^o)).
\end{align}
Considering the efficiency of matrix multiplication, we introduce \emph{im2col} operation to convert $X^{h,a}$ and $X^{h,m}$ to tensors sized $s^3\times H^l W^l D^l M^l$. In this way, we can multiply them with $X^w$ using Hadamard product and apply summation to aggregate the contribution of adjacent filters. Formally, the computation process can be formalized as:
\begin{align}
	X^p &= {{sum}}(X^w \odot {{im2col}}\ (X^{h,a}))\\
	X^q &= {{sum}}(X^w \odot {{im2col}}\ (X^{h,m})),
\end{align}
where $\odot$ denotes Hadamard product and $\{X^p,X^q\} \in \mathcal{R}^{H^l\times W^l\times D^l\times M^l}$. \emph{sum} is applied to the first dimension. Finally, we concatenate three outputs to form next layer's input $X^{l+1}$:
\begin{align}
	X^{l+1} = {concat}(X^{h,t}, X^{p}, X^{q})
\end{align}

Generally speaking, the reasoning module learns how to align the semantic representations of anchor and target volumes from seen organs. During the inference stage, the learned rule can be well applied to unseen classes using one-shot pair as supervision signals.

\subsubsection{Training and Inference} During the training stage, we first build a pool of annotated images for each organ class. In each training iteration, we randomly pick anchor and target samples from the same class pool (thus it is a binary segmentation task). The anchor input is fed to the top two encoders together with its annotation. Meanwhile, the target image is passed to the bottom encoder after registration, and its annotation is used as the ground truth for training. In practice, we use image batches as inputs considering the training efficiency. Particularly, we manually make each batch to have different class annotations which help to produce better segmentation results in our experiments. We train OrganNet with a combination of Dice and cross entropy losses with equal weights. With the large training pool, the organ concept is learned under full supervision. In the inference phase, when a previously unseen abdominal organ needs to be segmented, only one anchor image and its annotation are needed. It is worth noting that we pick the most similar example to the anatomical average computed for each organ class as the anchor image during the inference stage following the instruction from \cite{zhao2019data}.

\section{Experiments and Results}
In this section, we conduct experiments together with ablation studies to demonstrate the strength of OrganNet. First, we briefly introduce the dataset and evaluation metric used for experiments. Then, we present the implementation details of OrganNet and display the experimental results under various settings.

\subsection{Dataset and Evaluation Metric}\label{sec:data_eval}
We evaluate our method on 90 abdominal CT images collected from two publicly available datasets: 43 subjects from The Cancer Image Archive (TCIA) Pancreas CT dataset \cite{clark2013cancer} and 47 subjects from the Beyond the Cranial Vault (BTCV) Abdomen dataset \cite{landman2015miccai} with segmentations of 14 classes which include spleen, left/right kidneys, gallbladder, esophagus, liver, stomach, aorta, inferior vena cava, portal vein and splenic vein, pancreas, and left/right adrenal glands\footnote{\url{https://zenodo.org/record/1169361}.}. In practice, we test the effectiveness of the OrganNet on 5 kinds of unseen abdominal organs (spleen, right kidney, aorta, pancreas and stomach), which present great challenges because of their variations in size and shape, and use the rest 9 organs for training. We employ the Dice coefficient as our evaluation metric.

\subsection{Implementation Details} \label{sec:impl}
We build OrganNet based on 3D U-Net. To be specific, each encoder branch and the decoder (cf. Fig. \ref{framework}) share the same architecture as those of 3D U-Net. The initial channel number of the encoder is 8 which is doubled after features maps are downsampled. Moreover, as mentioned in Sec. \ref{sec:method} and Fig. \ref{framework}, we use different neighbour sizes in PRMs, which are $7\times 7\times 7$, $5\times 5\times 5$ and $3\times 3\times 3$ from shallow layers to deep layers, respectively. Following \cite{zhao2019data}, for each organ class, we pick the most similar example to the anatomical average computed for each organ class from the test set, and treat it as the anchor image during the inference stage. Moreover, we repeat each experiments for three times and report their standard deviation. We conduct all experiments with the PyTorch framework \cite{paszke2019pytorch}. We use the Adam optimizer \cite{kingma2014adam} and train our model for fifty epochs. The batch size is 8 (one per GPU) with 8 NVIDIA GTX 1080 Ti GPUs. The cosine annealing technique \cite{loshchilov2016sgdr} is adopted to adjust the learning rate from $10^{-3}$ to $10^{-6}$ with a weight decay of $10^{-4}$. For organs used to train the OrganNet, we use 80\% of data for training, and the remaining 20\% are used for validation. For organs used to test the OrganNet, we randomly select 20\% data for evaluation. The other 80\% are used to train a fully-supervised 3D U-Net.

\begin{table*}[!t]
	\centering
	\scriptsize
	\setlength{\belowcaptionskip}{-5pt}
	\begin{tabular}{ccccccc}
		\toprule
		\multirow{2}{*}{Method} & \multicolumn{6}{c}{Dice}        \\ \cline{2-7}
		& Spleen & Right Kidney & Aorta & Pancreas & Stomach & Mean   \\
		\midrule
		DataAug & 69.9$\scriptstyle{\pm{0.8}}$ & 73.6$\scriptstyle{\pm{0.7}}$ & 45.2$\scriptstyle{\pm{0.9}}$ & 48.8$\scriptstyle{\pm{0.9}}$ & 55.1$\scriptstyle{\pm{0.8}}$ & 58.5$\scriptstyle{\pm{0.8}}$\\
		Squeeze \& Excitation & 70.6$\scriptstyle{\pm{0.6}}$ &\ 75.2 $\scriptstyle{\pm{0.4}}$ & 47.7$\scriptstyle{\pm{0.6}}$ & 52.3$\scriptstyle{\pm{0.7}}$ & 57.5$\scriptstyle{\pm{0.8}}$ &\ 60.7 $\scriptstyle{\pm{0.6}}$\\
		LT-Net & 73.5$\scriptstyle{\pm{0.7}}$ & 76.4$\scriptstyle{\pm{0.5}}$ & 52.1$\scriptstyle{\pm{0.5}}$ & 54.8$\scriptstyle{\pm{0.6}}$ & 60.2$\scriptstyle{\pm{0.7}}$ & 63.4$\scriptstyle{\pm{0.6}}$\\
		\midrule 
		DataAug* & 79.6$\scriptstyle{\pm{0.6}}$ & 78.7$\scriptstyle{\pm{0.6}}$ & 63.8$\scriptstyle{\pm{0.7}}$ & 64.2$\scriptstyle{\pm{0.8}}$ & 77.3$\scriptstyle{\pm{0.6}}$ & 72.7$\scriptstyle{\pm{0.7}}$\\
		Squeeze \& Excitation* & 79.2$\scriptstyle{\pm{0.5}}$ & 80.7$\scriptstyle{\pm{0.4}}$ &\ 67.4 $\scriptstyle{\pm{0.7}}$ &  67.8$\scriptstyle{\pm{0.6}}$ & 78.1$\scriptstyle{\pm{0.5}}$ & 74.6$\scriptstyle{\pm{0.5}}$\\
		LT-Net* & 82.6$\scriptstyle{\pm{0.5}}$ & 82.9$\scriptstyle{\pm{0.5}}$ & 72.2$\scriptstyle{\pm{0.6}}$ & 70.7$\scriptstyle{\pm{0.5}}$ & 80.2 $\scriptstyle{\pm{0.4}}$ & 77.7$\scriptstyle{\pm{0.5}}$\\
		\midrule
		OrganNet & \textbf{89.1}$\scriptstyle{\pm{0.6}}$ & \textbf{86.0}$\scriptstyle{\pm{0.6}}$ & \textbf{77.0}$\scriptstyle{\pm{0.7}}$ & \textbf{72.8}$\scriptstyle{\pm{0.5}}$ & \textbf{82.6}$\scriptstyle{\pm{0.7}}$ & \textbf{81.5}$\scriptstyle{\pm{0.6}}$\\
		\bottomrule
	\end{tabular}
	\caption{Comparing OrganNet with DataAug \cite{zhao2019data}, Squeeze \& Excitation \cite{roy2020squeeze} and LT-Net \cite{wang2020lt}. ``*'' denotes that we \textbf{pretrain the segmentation network with 9 seen organ classes so that all approaches (including OrganNet) access the same set of labeled data which helps to guarantee the fairness of comparison.}}
	\label{dataaug}
\end{table*}
\begin{table}[!t]
	\centering
	\scriptsize
	\setlength{\belowcaptionskip}{-5pt}
	\begin{tabular}{ccc}
		\toprule
		Method & \tabincell{c}{Number of Training Samples \\for each unseen class} &Mean Dice \\
		\midrule
		3D U-Net & 1 & 20.9 $\scriptstyle{\pm{11.8}}$\\
		3D U-Net & 30\% & 79.3$\scriptstyle{\pm{0.4}}$\\ 
		\midrule
		OrganNet & 1 & 81.5$\scriptstyle{\pm{0.6}}$\\
		\midrule
		3D U-Net* & 100\% & 87.4$\scriptstyle{\pm{0.3}}$ \\
		\bottomrule
	\end{tabular}
	\caption{Comparison with 3D U-Net trained with different number of unseen class samples. ``*'' denotes the fully-supervised upper bound.}
	\label{fully_supervised}
\end{table}


\subsection{Comparison with One-shot Segmentation Methods: Better Performance} 
\label{sec:sota}
We compare our OrganNet against DataAug \cite{zhao2019data}, Squeeze \& Excitation \cite{roy2020squeeze} and LT-Net \cite{wang2020lt}. We do not compare with CTN \cite{lu2020learning} as it is based on human intervention. As we have mentioned above, a prerequisite of using atlas-based method is that the differences between anchor and target inputs should be small enough to learn appropriate transformation matrix. To enable DataAug and LT-Net to segment all 5 test organs, we propose two settings. The first setting is to use the original implementations which are based on only one annotated sample for each class. The second setting is to pretrain DataAug and LT-Net using 9 seen organ classes and then retrain \emph{5 independent} models using a number of each unseen class (denoted as * in Table \ref{dataaug}). In contrast, our OrganNet only needs one network to be trained once to segment all 5 organs. We report the results in Table \ref{dataaug}, from which we observe that our OrganNet outperforms the naive DataAug, Squeeze \& Excitation and LT-Net (without ``*'') by a significant margin. Even after these three models are pretrained using the other 9 classes' data, OrganNet is still able to surpass them by at least 3.8 percents. We believe the poor performance of DataAug and LT-Net may be explained by the fact that explicit transformation functions are difficult to learn in abdominal CT images, where large displacements usually happen. For Squeeze \& Excitation, we believe it may achieve better performance after adding its human-in-the-loop mechanism.


\subsection{Comparison with Supervised 3D U-Nets: Less Labeling Cost} 
\label{sec:fully_supervised}
Lastly, we compare our OrganNet with 3D U-Net \cite{cciccek20163d} which follows a supervised training manner. For the sake of fairness, we first pretrain 3D U-Net on 9 seen organs which are used to train OrganNet, and then fine-tune the 3D U-Net for each unseen class using different amount of labeled samples. Results are displayed in Table \ref{fully_supervised}.

One obvious observation is that using one training sample for unseen classes is far not enough for supervised baseline because the supervised training process may lead to severe overfitting problem. Even if we add more labeled samples to 30\%, our OrganNet can still achieve competitive results compared with supervised baseline. It is worth noting that annotating CT volumes is an intensive work which may cost several days of several well-trained radiologists. Thus, our OrganNet can greatly reduce the annotation cost. Finally, we offer a fully-supervised model which utilizes all training samples of unseen abdominal organs.

\begin{figure}[t]
	\begin{minipage}[b]{.45\columnwidth}
		\centering
		\setlength{\belowcaptionskip}{-5pt}
		\includegraphics[width=1.0\columnwidth]{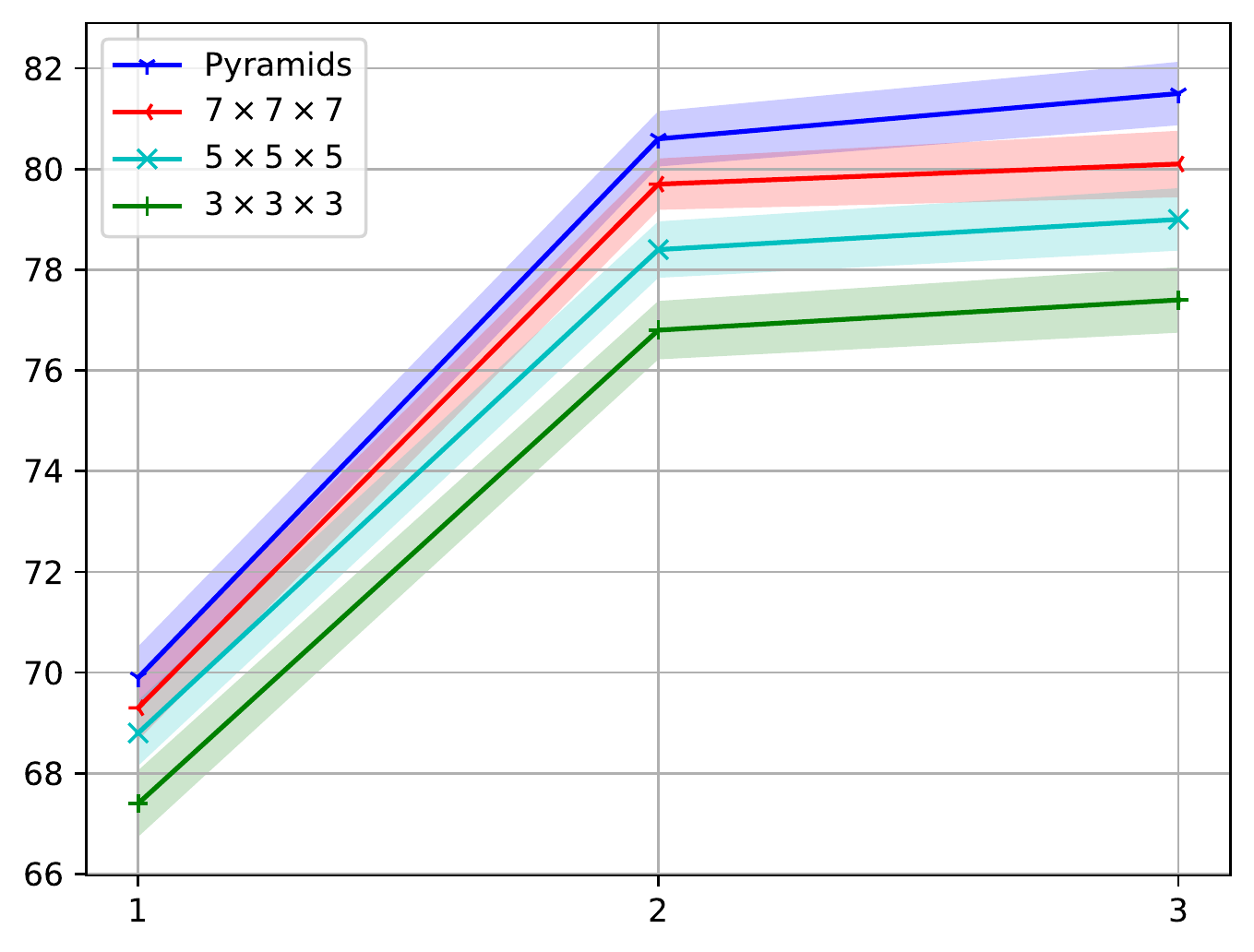}
		\caption{Mean Dice scores for varying number of reasoning modules  and different sizes of neighbour regions. The shaded region represents the standard deviation of each experiment setting.}
		\label{ab2}
	\end{minipage}%
	\hfill
	\begin{minipage}[b]{.45\columnwidth}
		\centering
		\setlength{\belowcaptionskip}{-5pt}
		\begin{tabular}{ll}
			\toprule
			Method & Mean Dice (\%)  \\
			\midrule
			{[}1, 2, 3{]} & 79.6$\scriptstyle{\pm{0.5}}$\\
			{[}1, 2{]} + 3 & 80.1$\scriptstyle{\pm{0.7}}$\\
			1 + 2 + 3 &  81.2 $\scriptstyle{\pm{1.2}}$\\
			\midrule
			1 + {[}2, 3{]} & 81.5 $\scriptstyle{\pm{0.6}}$\\
			\bottomrule
		\end{tabular}
		\captionof{table}{Influence of employing weight sharing across different encoders. For simplicity, we denote the encoder for target volume as \textbf{1}, the encoder for anchor volume as \textbf{2} and the encoder for anchor mask as \textbf{3}, respectively. Bracketed numbers mean that these encoders share the same model weights.}
		\label{ab1}
	\end{minipage}
\end{figure}


\subsection{Ablation Study on Network Design} 
\label{sec:ablation_study}

\subsubsection{Number of reasoning modules  and different sizes of neighbour regions.} In Fig. \ref{ab2}, we display the experimental results of using different numbers of PRMs and sizes of neighbour regions. We can find that adding more reasoning modules can consistently improve the model results and different sizes of neighbour regions behave similarly. If we compare the results of using different sizes of neighbour regions, it is easy to find that the proposed pyramid strategies works the best (red curve), even surpassing using $7\times 7\times 7$ which considers more adjacent regions. Such comparison verifies our hypothesis that deep layers require smaller neighbour sizes because their receptive fields are much larger while large kernel sizes may import additional noise. Interestingly, we can find that increasing the number from 2 to 3 would bring obvious improvements, suggesting adding more PRMs may not benefit a lot.

\subsubsection{Shared encoders or not?} Since the encoder of normal 3D U-Net is heavy, we study if it is possible to perform weight sharing across three different encoders. In Table \ref{ab1}, we report the experimental results of using different weight sharing strategies. It is obvious that sharing weights across all three encoders performs the worst. This phenomenon implies that different inputs may need different encoders. When making the bottom encoder (1) and the middle encoder (2) share weights, the average performance is slightly improve by approximate 0.5 percent. Somewhat surprisingly, building independent encoders (1+2+3) helps to improve the overall performance a lot, showing that learning specific features for each input is workable. Finally, when the top encoder shares the same weights with the middle encoder, our OrganNet is able to achieve the best average performance. Such results suggest that the top two encoders may complement each other.

\subsection{Visual Analysis}
\begin{figure}[t]
	\centering
	\setlength{\belowcaptionskip}{-5pt}
	\includegraphics[width=0.8\columnwidth]{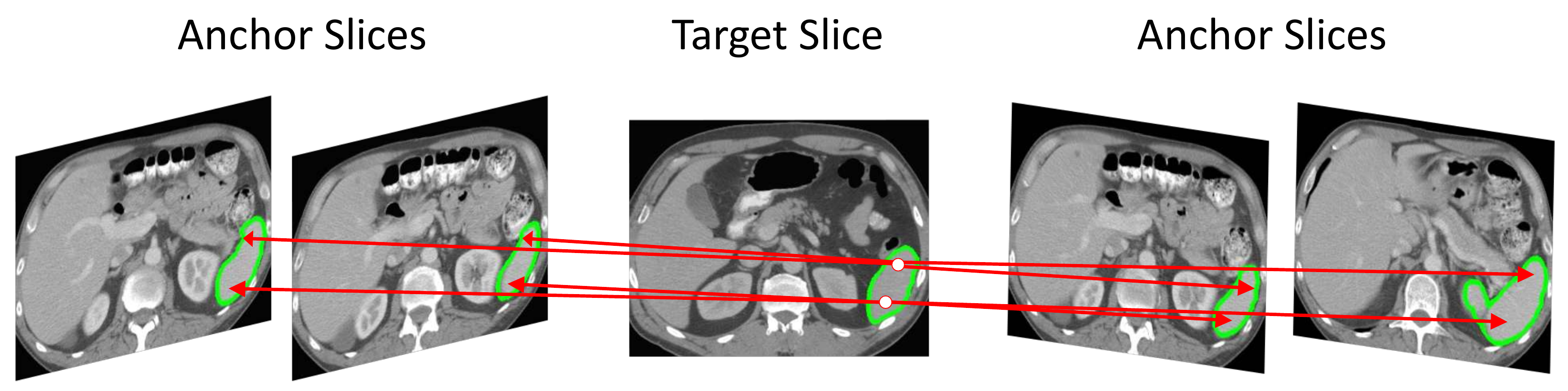}
	\caption{Given a specific position (white dot with a red border) in the target slice, we visualize the most related regions in adjacent slices (discovered by OrganNet) of the anchor image. The green contour denotes the spleen. The white dots in the target slice stand for two randomly selected positions.}
	\label{att}
\end{figure}
\begin{figure}[t]
	\centering
	\includegraphics[width=0.9\columnwidth]{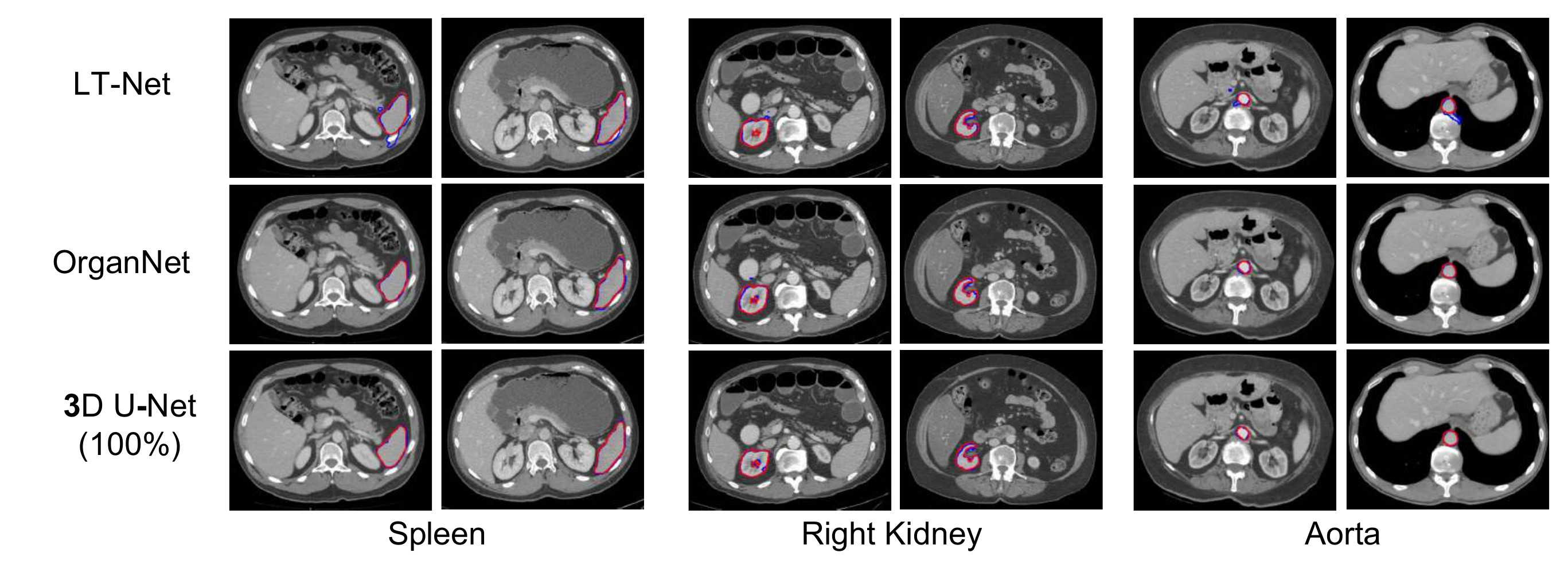}
	\caption{Visualization of segmentation results. The red contours denote the ground truths while the blue ones represent model predictions.}
	\label{vis}
\end{figure}
In this part, we conduct visual analyses on proposed OrganNet. Firstly, in Fig. \ref{att}, we visualize the most related regions learned by OrganNet. We can find that, given a specific position in the target slice, OrganNet can automatically discover most related regions in the anchor image based on computed anatomical similarity scores. In practice, OrganNet incorporates segmentation labels from these regions and produces a final prediction for the target region. From Fig. \ref{vis}, we can see that OrganNet is able to produce comparable results with 3D U-Net trained with 100\% labeled data.

\subsection{Generalization to Non-organ Segmentation}
\begin{table}[!t]
	\centering
	\scriptsize
	\setlength{\belowcaptionskip}{-8pt}
	\begin{tabular}{lcc}
		\toprule
		Method & \tabincell{c}{Number of Training Samples \\for Kidney Tumor} & Kidney Tumor Dice (\%) \\
		\midrule
		3D U-Net & 30\% & 77.1$\scriptstyle{\pm{0.9}}$\\ 
		OrganNet & 1 & 78.5$\scriptstyle{\pm{1.2}}$\\
		\midrule
		3D U-Net* & 100\% & 84.5$\scriptstyle{\pm{0.6}}$ \\
		\bottomrule
	\end{tabular}
	\caption{Comparison of OrganNet and 3D U-Net in tumor segmentation, both of which are pretrained with LiTS. ``*'' stands for the fully-supervised upper bound.}
	\label{tumor}
\end{table}
To better demonstrate the effectiveness of OrganNet, we conduct experiments on LiTS \cite{bilic2019liver} and KiTS \cite{heller2019kits19}, where we use LiTS to train OrganNet and test it on KiTS with only one labeled sample (following the one-shot segmentation scenario. From Table \ref{tumor}, we can see that OrganNet still maintains its advantages over 3D U-Net trained with 30\% labeled data, demonstrating the generalization ability of OrganNet to non-organ segmentation problems.

\section{Conclusions and Future Work}
In this paper, we present a novel one-shot medical segmentation approach that enables us to learn a generalized organ concept by employing a reasoning process between anchor and target CT volumes. The proposed OrganNet models the key components, i.e., the anatomical similarity between images, within a single deep learning framework. Extensive experiments demonstrate the effectiveness of proposed OrganNet.

%
%

\bibliographystyle{splncs04}
\bibliography{mybibliography}

\begin{thebibliography}{10}
\providecommand{\url}[1]{\texttt{#1}}
\providecommand{\urlprefix}{URL }
\providecommand{\doi}[1]{https://doi.org/#1}

\bibitem{bentaieb2016topology}
BenTaieb, A., Hamarneh, G.: Topology aware fully convolutional networks for
  histology gland segmentation. In: MICCAI. pp. 460--468 (2016)

\bibitem{bertinetto2016fully}
Bertinetto, L., Valmadre, J., Henriques, J.F., Vedaldi, A., Torr, P.H.:
  {Fully-Convolutional {S}iamese Networks for Object Tracking}. In: ECCV. pp.
  850--865 (2016)

\bibitem{bilic2019liver}
Bilic, P., Christ, P.F., et~al.: {The Liver Tumor Segmentation Benchmark
  (LiTS)}. arXiv preprint arXiv:1901.04056  (2019)

\bibitem{cciccek20163d}
{\c{C}}i{\c{c}}ek, {\"O}., Abdulkadir, A., Lienkamp, S.S., Brox, T.,
  Ronneberger, O.: {3D U-Net: Learning Dense Volumetric Segmentation from
  Sparse Annotation}. In: MICCAI. pp. 424--432 (2016)

\bibitem{clark2013cancer}
Clark, K., Vendt, B., Smith, K., et~al.: {The Cancer Imaging Archive ({TCIA}):
  {M}aintaining and Operating a Public Information Repository}. Journal of
  Digital Imaging  \textbf{26}(6),  1045--1057 (2013)

\bibitem{dalca2019unsupervised}
Dalca, A.V., Yu, E., Golland, P., Fischl, B., Sabuncu, M.R., Iglesias, J.E.:
  {Unsupervised Deep Learning for Bayesian Brain MRI Segmentation}. In: MICCAI.
  pp. 356--365. Springer (2019)

\bibitem{dinsdale2019spatial}
Dinsdale, N.K., Jenkinson, M., Namburete, A.I.: {Spatial Warping Network for
  {3D} Segmentation of the Hippocampus in {MR} Images}. In: MICCAI. pp.
  284--291 (2019)

\bibitem{gentner1997reasoning}
Gentner, D., Holyoak, K.J.: {Reasoning and Learning by Analogy: Introduction}.
  American Psychologist  \textbf{52}(1), ~32 (1997)

\bibitem{gibson2018automatic}
{Gibson}, E., {Giganti}, F., {Hu}, Y., {Bonmati}, E., {Bandula}, S.,
  {Gurusamy}, K., {Davidson}, B., {Pereira}, S.P., {Clarkson}, M.J., {Barratt},
  D.C.: {Automatic Multi-Organ Segmentation on Abdominal CT With Dense
  V-Networks}. IEEE Transactions on Medical Imaging  \textbf{37}(8),
  1822--1834 (2018)

\bibitem{heinrich2013mrf}
Heinrich, M.P., Jenkinson, M., Brady, M., Schnabel, J.A.: {MRF-based Deformable
  Registration and Ventilation Estimation of Lung CT}. IEEE Transactions on
  Medical Imaging  \textbf{32}(7),  1239--1248 (2013)

\bibitem{heller2019kits19}
Heller, N., Sathianathen, N., Kalapara, A., et~al.: {The KiTS19 Challenge Data:
  300 Kidney Tumor Cases with Clinical Context, CT Semantic Segmentations, and
  Surgical Outcomes}. arXiv preprint arXiv:1904.00445  (2019)

\bibitem{iglesias2015multi}
Iglesias, J.E., Sabuncu, M.R.: {Multi-Atlas Segmentation of Biomedical Images:
  a Survey}. Medical image analysis  \textbf{24}(1),  205--219 (2015)

\bibitem{kingma2014adam}
Kingma, D.P., Ba, J.: {Adam: {A} Method for Stochastic Optimization}. arXiv
  preprint arXiv:1412.6980  (2014)

\bibitem{landman2015miccai}
Landman, B., Xu, Z., Eugenio~Igelsias, J., et~al.: {MICCAI Multi-Atlas Labeling
  Beyond the Cranial Vault--Workshop and Challenge} (2015)

\bibitem{liang2019comparenet}
Liang, Y., Song, W., Dym, J., Wang, K., He, L.: {CompareNet: Anatomical
  Segmentation Network with Deep Non-local Label Fusion}. In: MICCAI. pp.
  292--300 (2019)

\bibitem{loshchilov2016sgdr}
Loshchilov, I., Hutter, F.: {SGDR: Stochastic Gradient Descent with Warm
  Restarts}. arXiv preprint arXiv:1608.03983  (2016)

\bibitem{lu2020learning}
Lu, Y., Zheng, K., Li, W., Wang, Y., Harrison, A.P., Lin, C., Wang, S., Xiao,
  J., Lu, L., Kuo, C.F., et~al.: {Learning to Segment Anatomical Structures
  Accurately from One Exemplar}. MICCAI  (2020)

\bibitem{paszke2019pytorch}
Paszke, A., Gross, S., Massa, F., et~al.: {PyTorch: an Imperative Style,
  High-Performance Deep Learning Library}. In: NeurIPS. pp. 8024--8035 (2019)

\bibitem{ravishankar2017joint}
Ravishankar, H., Thiruvenkadam, S., Venkataramani, R., Vaidya, V.: {Joint Deep
  Learning of Foreground, Background and Shape for Robust Contextual
  Segmentation}. In: IPMI. pp. 622--632 (2017)

\bibitem{ravishankar2017learning}
Ravishankar, H., Venkataramani, R., Thiruvenkadam, S., Sudhakar, P., Vaidya,
  V.: {Learning and Incorporating Shape Models for Semantic Segmentation}. In:
  MICCAI. pp. 203--211 (2017)

\bibitem{roy2020squeeze}
Roy, A.G., Siddiqui, S., P{\"o}lsterl, S., Navab, N., Wachinger, C.:
  {‘Squeeze \& Excite’ Guided Few-shot Segmentation of Volumetric Images}.
  Medical Image Analysis  \textbf{59},  101587 (2020)

\bibitem{trullo2019multiorgan}
Trullo, R., Petitjean, C., Dubray, B., Ruan, S.: {Multiorgan Segmentation using
  Distance-Aware Adversarial Networks}. Journal of Medical Imaging
  \textbf{6}(1),  014001 (2019)

\bibitem{wang2020lt}
Wang, S., Cao, S., Wei, D., Wang, R., Ma, K., Wang, L., Meng, D., Zheng, Y.:
  {LT-Net: Label Transfer by Learning Reversible Voxel-wise Correspondence for
  One-shot Medical Image Segmentation}. In: CVPR. pp. 9162--9171 (2020)

\bibitem{xu2016evaluation}
Xu, Z., Lee, C.P., Heinrich, M.P., Modat, M., Rueckert, D., Ourselin, S.,
  Abramson, R.G., Landman, B.A.: {Evaluation of Six Registration Methods for
  the Human Abdomen on Clinically Acquired CT}. IEEE Transactions on Biomedical
  Engineering  \textbf{63}(8),  1563--1572 (2016)

\bibitem{zhao2019data}
Zhao, A., Balakrishnan, G., Durand, F., Guttag, J.V., Dalca, A.V.: {Data
  Augmentation using Learned Transformations for One-Shot Medical Image
  Segmentation}. In: CVPR. pp. 8543--8553 (2019)

\bibitem{zhou2017sunrise}
Zhou, H.Y., Gao, B.B., Wu, J.: {Sunrise or Sunset: Selective Comparison
  Learning for Subtle Attribute Recognition}. arXiv preprint arXiv:1707.06335
  (2017)

\bibitem{zhou2018semi}
Zhou, H.Y., Oliver, A., Wu, J., Zheng, Y.: {When Semi-supervised Learning Meets
  Transfer Learning: Training strategies, Models and Datasets}. arXiv preprint
  arXiv:1812.05313  (2018)

\bibitem{zhou2020comparing}
Zhou, H.Y., Yu, S., Bian, C., Hu, Y., Ma, K., Zheng, Y.: {Comparing to learn:
  Surpassing imagenet pretraining on radiographs by comparing image
  representations}. In: MICCAI. pp. 398--407. Springer (2020)

\end{thebibliography}






\end{document}